\documentclass[sigconf,nonacm]{acmart}

\AtBeginDocument{%
  }

\setcopyright{none}
\usepackage{booktabs}
\usepackage{tabularx}
\usepackage{array}
\usepackage{amsmath}
\newcolumntype{Y}{>{\centering\arraybackslash}X}
\usepackage{colortbl}
\usepackage{bm}
\usepackage{graphicx}
\usepackage{subcaption}
\usepackage{float}
\usepackage[linesnumbered,ruled,vlined]{algorithm2e}

\usepackage{amssymb}
\usepackage[dvipsnames,table]{xcolor}
\definecolor{rose_red}{RGB}{219, 48, 122}
\definecolor{cite_blue}{RGB}{70, 105, 170}
\definecolor{pinkrowA}{RGB}{255, 250, 252}
\definecolor{pinkrowB}{RGB}{255, 235, 244}
\definecolor{pinkrowC}{RGB}{252, 215, 232}
\AtBeginDocument{%
  \hypersetup{linkcolor=cite_blue,citecolor=cite_blue,urlcolor=cite_blue}}

\begin{document}

\title{Holistic Reliability Propagation: Decoupling Annotation and Prediction for Robust Noisy-Label Learning}

\author{Jingyang Mao}
\affiliation{%
  \institution{Nanjing Normal University}
  \city{Nanjing}
  \country{China}}
\author{Ningkang Peng}
\affiliation{%
  \institution{Nanjing Normal University}
  \city{Nanjing}
  \country{China}}
\author{Yanhui Gu}
\affiliation{%
  \institution{Nanjing Normal University}
  \city{Nanjing}
  \country{China}}

% \affiliation{%

% \affiliation{%

% \affiliation{%

% \affiliation{%

% \affiliation{%

% \affiliation{%

% \affiliation{%

\renewcommand{\shortauthors}{Anonymous Author(s)}

\begin{abstract}
Learning with noisy labels in multimedia classification often combines external annotations and model predictions into a single reliability weight, even though the two sources can fail for different reasons. We instead estimate \emph{disentangled reliabilities}: bilevel meta-learning produces two batch-normalized scalars per sample, $\alpha$ for the given label and $\beta$ for the pseudo-label, without constraining them to sum to one. Holistic Reliability Propagation (HRP) then routes them to different objectives, using reliability-aware Mixup with global gating on the input branch and $\beta$-gated pseudo-label positives on the contrastive branch. On synthetic and real-world benchmarks, HRP improves average accuracy over strong baselines and remains competitive at the highest noise rates.
\end{abstract}


% \ccsdesc[500]{Computer systems organization~Neural networks}
% \keywords{Noisy-label learning, Reliability propagation, Contrastive learning}
\maketitle


\section{Introduction}

Deep neural networks are the default choice for multimedia classification, yet their supervision is frequently imperfect: crowd labels disagree with the image, categories are ambiguous \cite{northcutt2021confident}, and models must learn from a mixture of external annotations and their own predictions. In learning with noisy labels \cite{song2025survey}, these two cues are often combined in a single training signal. Representative pipelines either blend the given label with a pseudo-label \cite{li2020dividemix} or rely on one fused reliability weight. Such designs implicitly treat annotation quality and predictive confidence as interchangeable aspects of one reliability score. In practice, however, the two can disagree: a visually easy instance may carry a wrong human label, while a hard instance may have a correct label but an overconfident wrong prediction. Collapsing both into one scalar makes it difficult to interpret whether a large loss reflects a bad label, an unreliable prediction, or both, and it encourages uniform reweighting that does not match how errors arise.

We further note that forcing the two cues into a zero-sum pair, such as weights that must sum to one \cite{zhou2024l2b}, is restrictive when neither source should strongly drive updates or when both are trustworthy and should remain so. Separate estimates of how useful the annotator's label is versus how useful the model's pseudo-label is allow both to be down-weighted together when neither is informative, without committing to a fixed trade-off. We denote these two batch-normalized, meta-learned emphases by $\alpha$ for the given label and $\beta$ for the pseudo-label. The remaining question is how to use them in training, because not every objective should consume $\alpha$ and $\beta$ in the same way.

Input-level regularization and representation learning answer different questions. Mixup-style augmentation \cite{mixup} must decide how strongly two instances should be interpolated and whether the resulting synthetic example should contribute gradients at all; for that, an overall sense of whether the pair jointly carries usable signal is natural, which we summarize as the clamped total reliability $r_i=\mathrm{Clamp}(\alpha_i+\beta_i,r_{\min},r_{\max})$. In contrast, contrastive objectives \cite{chen2020simple} define positive pairs in feature space. Supervised and noise-robust variants further adapt this idea under class-aware or reliability-aware supervision \cite{supervised,li2021rrl,unicon}. If positives were tied directly to noisy given labels, erroneous annotations would pull disparate visuals together and damage the embedding geometry. We therefore keep pseudo-label-based positives for the contrastive branch and gate pairwise attraction with normalized $\beta_i\beta_j$, so that attraction is weakened when the model is not consistently confident about the pseudo-labels, without routing $\alpha$ into the definition of those positives.

Holistic Reliability Propagation (HRP) instantiates this split. Bilevel meta-learning on a small clean validation set produces $\alpha_i$ and $\beta_i$ from truncated, batch-normalized meta-gradients of two cross-entropy terms. At the input branch, Reliability-Arbitrated Mixup (RAM) uses $r_i$ and $r_j$ to parameterize an asymmetric Beta law for the Mixup coefficient, and Global Reliability Gating (GRG) scales each RAM term by $\max(r_i,r_j)$. At the representation branch, Consensus-Driven Contrastive Learning (CDCL) applies $\beta$-gated pseudo-label positives as described above. Two networks co-train by exchanging targets to mitigate confirmation bias \cite{Han9}.

The main contributions of this paper are summarized as follows:
\begin{itemize}
    \item We center noisy-label learning on \emph{disentangled reliabilities} for external labels ($\alpha$) and pseudo-labels ($\beta$), estimated with bilevel meta-gradients instead of a single mixing coefficient that forces a zero-sum trade-off.
    \item We instantiate this principle in HRP: RAM with asymmetric Beta Mixup and GRG for the input branch, CDCL with $\beta$-gated positives for the representation branch, and dual networks exchanging targets.
    \item On the main CIFAR and Animal-10N benchmarks, HRP achieves the best average classification accuracy and improves AUROC in several noisy OOD settings.
\end{itemize}

\begin{figure*}[t]
    \centering
    \includegraphics[width=0.99\textwidth]{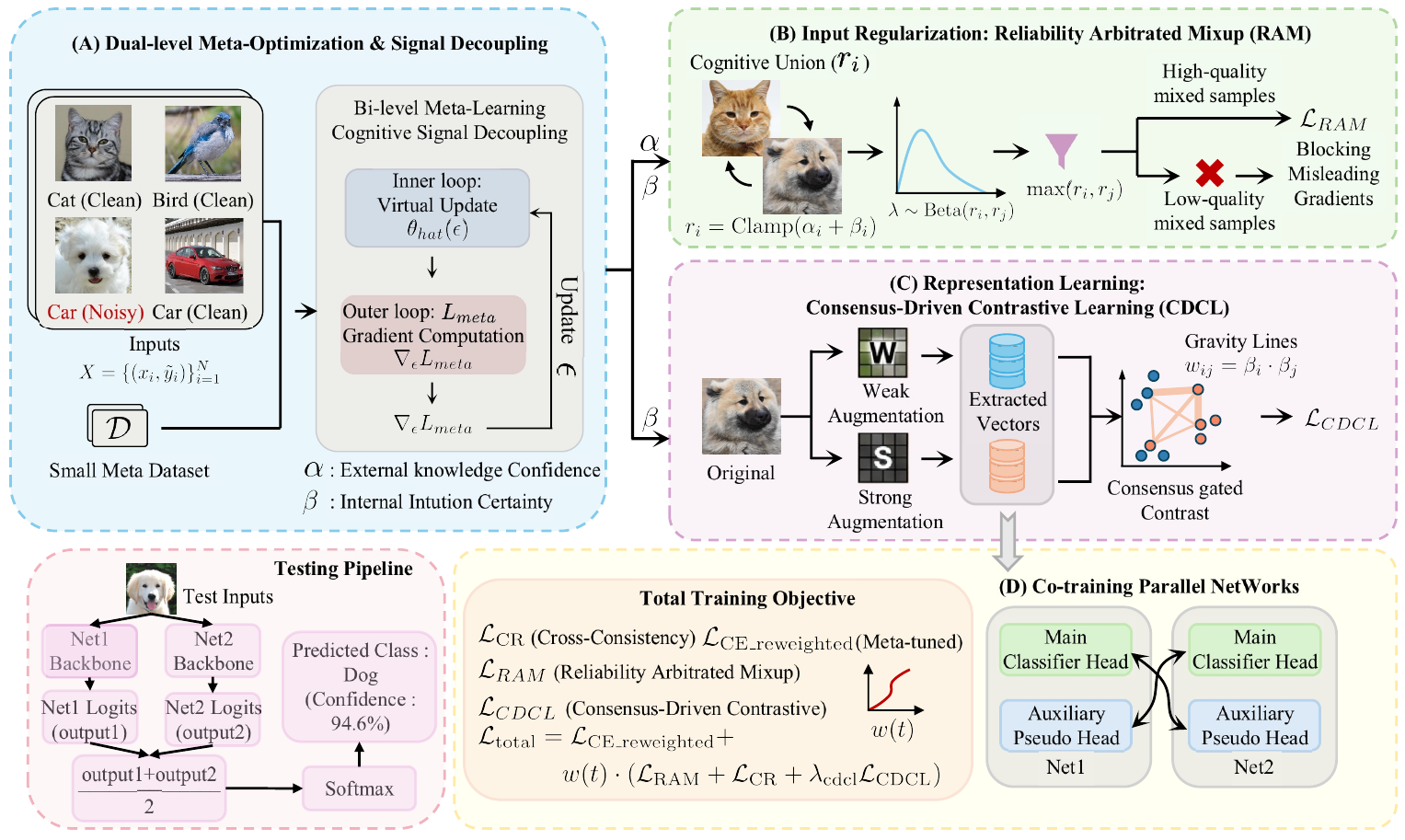}
    \Description{Block diagram of the HRP framework. Bilevel meta-learning estimates per-sample weights alpha for the given label and beta for the pseudo-label. RAM applies asymmetric Beta Mixup and global reliability gating on the input branch; CDCL applies beta-gated pseudo-label positives on the representation branch. Two networks co-train and exchange targets.}
    \caption{Overview of HRP: $\alpha$ and $\beta$ denote the reliabilities of the given label and the pseudo-label, respectively. The figure summarizes bilevel estimation, RAM with GRG, and CDCL with $\beta$-gated pseudo-label positives; for readability, the RAM box uses schematic notation, while the exact clamping and Beta parameterization are given in the method section.}
    \label{fig:1}
\end{figure*}
\section{Related Work}

\subsection{Sample Selection and Semi-Supervised Learning}
In the field of Learning with Noisy Labels (LNL), sample selection based on the memorization effect combined with Semi-Supervised Learning (SSL) is currently the most dominant paradigm. Deep neural networks typically prioritize fitting simple, clean patterns before gradually memorizing random noisy labels \cite{arpit2017closer,bai2021understanding}. Based on this phenomenon, the small-loss criterion is widely used to separate clean samples from noisy ones \cite{Gui13}. Early classic studies, such as Co-teaching and Co-teaching+, utilize a dual-network architecture to cross-train each other with low-loss samples, effectively alleviating the confirmation bias that a single network easily falls into during sample selection \cite{Han9,yu2019does}. Beyond sample selection, another line of work estimates transition structure or corrects losses directly \cite{patrini2017making,goldberger2017training,unsupervised}, while more recent studies emphasize instance-dependent corruption and label uncertainty \cite{xia2020part,northcutt2021confident}. Building upon this theoretical foundation, the deep integration of sample division with powerful semi-supervised learning techniques has pushed LNL tasks to new heights. A representative work, DivideMix \cite{li2020dividemix}, innovatively fits a two-component Gaussian Mixture Model (GMM) on the cross-entropy loss of each sample to calculate the posterior probability of a sample belonging to the clean distribution. This dynamically divides the training data into a labeled clean subset and an unlabeled noisy subset, followed by the application of MixMatch \cite{mixmatch} for semi-supervised training, establishing the mainstream pipeline in this field. Subsequent research has been dedicated to further breaking through the performance bottlenecks of sample division under extreme noise, primarily focusing on optimizing the precision and recall of the division. For instance, LongReMix \cite{longremix} points out that GMM fitting is prone to failure under extremely high noise rates, and thus proposes a two-stage algorithm. Complementary to this, ProMix \cite{promix} observes that overly strict filtering criteria often omit a large number of clean samples containing rich semantic information. Therefore, it attempts to maximize the utility of clean samples by proposing a matched high-confidence selection technique, which dynamically expands the base clean set by aligning predictive probabilities with given labels, thereby significantly raising the lower bound of representation learning in the SSL stage.

\subsection{Robust Representation and Contrastive Learning}
In multimedia analysis tasks, the visual feature space of images is typically extremely complex. Therefore, learning a robust underlying feature representation is often more critical than merely correcting classification labels at the top layer. Contrastive Learning (CL) demonstrates powerful representation capabilities by pulling positive sample pairs closer and pushing negative sample pairs apart in the latent space. Early self-supervised contrastive learning naturally shields label information by constructing positive pairs solely from different augmented views of the same image, thereby exhibiting strong robustness to label noise \cite{he2020momentum}. However, lacking the guidance of category labels, self-supervised CL is unable to form compact cluster structures. Subsequently, supervised contrastive learning (SupCon) was proposed \cite{supervised}, which leverages category labels to pull different instances of the same category closer together. Unfortunately, in LNL scenarios, directly applying SupCon encounters catastrophic failure: erroneous labels force the network to aggregate multimedia images with completely different visual semantics in the feature space, directly triggering feature collapse. Related work therefore explores selective contrastive positives, graph-based label cleaning, and semi-supervised contrastive refinement under noisy supervision \cite{Sel15,zhong2019graph,manifoldDivideMix,scanmix}. To harness the representation ability of contrastive learning while resisting the intrusion of label noise, recent works have begun to couple sample selection with contrastive learning. For instance, Robust Representation Learning (RRL) \cite{li2021rrl} mitigates representation degradation by embedding images into a low-dimensional subspace and regularizing its geometric structure using a contrastive mechanism that includes an unsupervised consistency loss. UNICON \cite{unicon} further observes that traditional small-loss sample selection disproportionately favors easy categories, leading to class imbalance in the representation space, and combines uniform selection with contrastive learning to address extreme label noise.

\subsection{Data Augmentation and Meta-Learning}
In multimedia representation learning, data augmentation has been widely used to smooth the decision boundaries of the feature space and prevent model overfitting. However, under extreme noisy label environments, the effectiveness of standard Mixup is severely challenged. Traditional Mixup relies on a symmetric Beta distribution, blindly performing proportional linear interpolation on pairs of images and their labels. This indiscriminate mechanism easily triggers the problem of manifold intrusion in multimedia data: symmetrically mixing clean, high-information features with noisy features containing severe annotation errors generates synthesized samples that suffer from deep ambiguity both visually and semantically, further polluting the originally clear feature manifold. Orthogonal efforts curb memorization with robust losses or early-stage regularization \cite{GCE,liu2020early}, revisit interpolation itself for noisy supervision \cite{MOIT14}, or explicitly model feature-dependent label noise instead of assuming a single global corruption pattern \cite{PLC}. To achieve dynamic soft evaluation of samples without relying on heuristic, hard loss thresholds, meta-learning has been introduced into the optimization trajectory of denoising tasks. Meta-learning methods typically utilize a small set of clean validation data as a meta-prior to guide the training of the main network. For example, MWNet \cite{shu2019mwnet} learns an explicit meta-weighting function by constructing a Multi-Layer Perceptron (MLP), achieving an adaptive mapping from sample loss to sample weight and reducing the need for manual hyperparameter tuning. Recently, L2B \cite{zhou2024l2b} further allows the model to bootstrap using its own predictions; it dynamically calculates the importance weight between true observed labels and model-generated labels via bilevel meta-optimization, thereby alleviating the confirmation bias brought by erroneous pseudo-labels to a certain extent.

\section{Method}

\subsection{Problem Formulation and Overall Framework}

\subsubsection{Problem Formulation}
Consider a multimedia image classification task with $N$ samples, where the training dataset is defined as $\mathcal{D} = \{(x_i, y_i)\}_{i=1}^N$. Here, $x_i \in \mathcal{X}$ denotes the input image, and $y_i \in \{0, 1\}^C$ represents the potentially corrupted one-hot label provided by external annotators over $C$ classes. Due to the semantic complexity of multimedia data and the inherent unreliability of manual annotation, $y_i$ is not entirely equivalent to the underlying true label $y_i^*$.

In Learning with Noisy Labels (LNL), our goal is to train a highly robust deep neural network $f_\theta: \mathcal{X} \rightarrow \mathbb{R}^C$ capable of minimizing the generalization error on a clean test distribution, even without prior knowledge of $y_i^*$. Existing semi-supervised LNL methods typically rely on Gaussian Mixture Models (GMM) for sample division and widely adopt data augmentation techniques to smooth decision boundaries. However, traditional Mixup \cite{mixup} samples the interpolation coefficient from a symmetric distribution $\lambda \sim \text{Beta}(\gamma, \gamma)$. This symmetry imposes a restrictive assumption: the two mixed samples $x_i$ and $x_j$ possess equal semantic dominance. In noisy scenarios, unable to distinguish the information quality of samples, symmetric Mixup blindly interpolates clear visual features with meaningless erroneous labels, thereby polluting the original feature manifold and distorting the learned feature geometry.

\subsubsection{Overall Framework}
To address conflicting uses of labels and predictions, we use the \emph{Holistic Reliability Propagation} (HRP) framework: meta-learning supplies per-sample $\alpha_i$ and $\beta_i$, and different modules consume them according to their role. Training uses two networks (Net1 and Net2) that exchange pseudo-labels to limit confirmation bias. The pipeline has three parts. First, bilevel meta-optimization perturbs the cross-entropy weights for the given label and for the pseudo-label on each training mini-batch, evaluates on a small clean meta-set, and maps meta-gradients to non-negative, batch-normalized $\alpha_i$ and $\beta_i$. Second, Reliability-Arbitrated Mixup (RAM) sets $r_i=\mathrm{Clamp}(\alpha_i+\beta_i,r_{\min},r_{\max})$ and samples $\lambda \sim \mathrm{Beta}\big(\gamma \frac{r_i}{r_i+r_j+\delta},\, \gamma \frac{r_j}{r_i+r_j+\delta}\big)$ for Mixup; Global Reliability Gating (GRG) then applies $w_{mix}^{(i,j)}=\max(r_i,r_j)$ directly to each RAM loss term so that pairs with weak total reliability contribute less. Third, Consensus-Driven Contrastive Learning (CDCL) forms cross-view positives from pseudo-labels and weights each positive pair by normalized $\beta_i\beta_j$, without routing $\alpha$ into the contrastive positives. Figure~\ref{fig:1} summarizes the layout; its RAM box uses schematic shorthand, while the exact clamping and asymmetric Beta parameterization are defined below.

\subsection{Bilevel Meta-Learning for Reliability Signal Disentanglement}
In existing LNL literature, a mainstream paradigm is to fuse the given noisy label $y_i$ with the model-generated pseudo-label $\tilde{y}_i$ via a single mixing coefficient (e.g., $\hat{y} = \alpha y_i + (1-\alpha)\tilde{y}_i$) to rectify the training objective. Such formulations allocate trust with one scalar, so increasing reliance on one source necessarily decreases reliance on the other. In contrast, our formulation treats the utilities of the given label and the pseudo-label as two independent signals, denoted by $\alpha_i$ and $\beta_i$. This zero-sum trade-off becomes restrictive in complex real-world multimedia noise: when both sources are unreliable, the model should be able to set $\alpha_i \approx 0$ and $\beta_i \approx 0$ simultaneously; conversely, when both are helpful, both should remain high. We therefore utilize a bilevel meta-learning framework to explicitly disentangle these two reliability signals.

\subsubsection{Inner-Loop Virtual Update}
To independently quantify the contribution of these two reliability signals, we extract a minimal clean meta-dataset $\mathcal{D}_{meta} = \{(x_j^{v}, y_j^{v})\}_{j=1}^M$ from the validation set. For any sample $x_i$ in a training mini-batch $\mathcal{B}$, let $\hat{c}_i = \arg\max f_\theta(x_i)$ denote the predicted class index, and let $\tilde{y}_i \in \{0,1\}^C$ denote the corresponding one-hot pseudo-label. We bind two sample-level learnable perturbation parameters, $\epsilon_{1,i}$ and $\epsilon_{2,i}$ (both initialized to zero), to the cross-entropy losses of the external label $y_i$ and the pseudo-label $\tilde{y}_i$, respectively. The perturbed composite training loss $\mathcal{L}_{train}$ is defined as:
\begin{equation}
\mathcal{L}_{train}(\theta, \epsilon) = \frac{1}{|\mathcal{B}|} \sum_{i \in \mathcal{B}} \Big( \epsilon_{1,i} \mathcal{L}_{CE}(f_\theta(x_i), y_i) + \epsilon_{2,i} \mathcal{L}_{CE}(f_\theta(x_i), \tilde{y}_i) \Big)
\end{equation}
In the inner loop, we construct a local SGD optimizer for this mini-batch to perform a single-step virtual forward update on the model parameters $\theta$ using $\mathcal{L}_{train}$, yielding temporary parameters $\hat{\theta}(\epsilon)$:
\begin{equation}
\hat{\theta}(\epsilon) = \theta - \eta \nabla_\theta \mathcal{L}_{train}(\theta, \epsilon)
\end{equation}
where $\eta$ is the learning rate of the current primary network.

\subsubsection{Outer-Loop Meta-Objective}
Subsequently, in the outer loop, we evaluate the generalization performance of this temporary model $\hat{\theta}(\epsilon)$, driven by the perturbation parameter $\epsilon$, on the clean meta-dataset $\mathcal{D}_{meta}$. To prevent the meta-data from polluting the Batch Normalization statistics in the backbone network, we strictly freeze the network states (i.e., enable evaluation mode) during the meta-loss computation. The meta-optimization objective $\mathcal{L}_{meta}$ is formulated as:
\begin{equation}
\mathcal{L}_{meta}(\hat{\theta}(\epsilon)) = \frac{1}{M} \sum_{j=1}^M \mathcal{L}_{CE}(f_{\hat{\theta}(\epsilon)}(x_j^{v}), y_j^{v})
\end{equation}

\subsubsection{Reliability Signal Disentanglement}
According to the chain rule of meta-learning, the gradient of the meta-loss with respect to the perturbation parameter $\epsilon$, denoted as $\nabla_\epsilon \mathcal{L}_{meta}$, precisely reflects the impact of increasing the weight of $y_i$ or $\tilde{y}_i$ on the model's generalization error. To optimize the model towards reducing the validation error, we extract the negative values of these gradients. Meanwhile, to filter out negative signals that would cause performance degradation, we introduce a zero-truncation (Clamp) mechanism. To ensure numerical stability under extremely high noise rates, smooth normalization is performed along the batch dimension. Ultimately, the explicitly disentangled external-label reliability $\alpha_i$ and pseudo-label reliability $\beta_i$ are calculated as:
\begin{equation}
\tilde{w}_{1,i} = \max\left(-\frac{\partial \mathcal{L}_{meta}}{\partial \epsilon_{1,i}}\Big|_{\epsilon=0}, 0\right), \quad \tilde{w}_{2,i} = \max\left(-\frac{\partial \mathcal{L}_{meta}}{\partial \epsilon_{2,i}}\Big|_{\epsilon=0}, 0\right)
\end{equation}
\begin{equation}
\alpha_i = \frac{\tilde{w}_{1,i}}{\sum_{k \in \mathcal{B}} (\tilde{w}_{1,k} + \tilde{w}_{2,k}) + \xi} \times |\mathcal{B}|, \quad \beta_i = \frac{\tilde{w}_{2,i}}{\sum_{k \in \mathcal{B}} (\tilde{w}_{1,k} + \tilde{w}_{2,k}) + \xi} \times |\mathcal{B}|
\end{equation}
where $\xi = 10^{-10}$ is a minimal constant to prevent division by zero. These disentangled weights provide the reliability priors used by RAM in the input branch and by CDCL in the representation branch.

\subsection{Input Regularization: Reliability-Arbitrated Mixup (RAM)}
In multimedia data, data augmentation is widely utilized to smooth decision boundaries in the feature space and mitigate overfitting. However, under extreme noisy label environments, directly applying traditional data augmentation methods triggers severe feature manifold intrusion. To address this issue, we propose the \emph{Reliability-Arbitrated Mixup} (RAM) module, which transforms blind random mixing into reliability-aware directional augmentation.

\subsubsection{Asymmetric Beta Parameterization}
Traditional Mixup samples the interpolation coefficient $\lambda$ from a symmetric Beta distribution $\text{Beta}(\gamma, \gamma)$, which implicitly assumes that the two source samples should contribute equally on average. Under noisy supervision, this assumption is often too restrictive because one sample may carry much weaker label or feature information than the other. RAM therefore uses the disentangled reliabilities from Section 3.2 to reshape the Beta distribution. We first define the sample's \emph{Total Reliability}:
\begin{equation}
r_i = \text{Clamp}(\alpha_i + \beta_i, r_{\min}, r_{\max})
\end{equation}
where $r_{\min}$ and $r_{\max}$ are truncation boundaries used to prevent numerical instability. When mixing sample $x_i$ and $x_j$, RAM uses the normalized total reliability $r$ to dynamically parameterize an asymmetric Beta distribution:
\begin{equation}
\lambda \sim \text{Beta}\left( \gamma \frac{r_i}{r_i + r_j + \delta}, \ \gamma \frac{r_j}{r_i + r_j + \delta} \right)
\end{equation}
where $\gamma$ is the baseline concentration parameter controlling the distribution width, and $\delta$ is a tiny constant preventing division by zero errors. In this way, the interpolation direction depends on the relative reliability of the two samples rather than on a fixed symmetric prior.

\subsubsection{Boundary Dynamics in Three Scenarios}
We summarize three limiting behaviors of the RAM Beta parameterization. When $r_i$ and $r_j$ are both large and similar, the two Beta shape parameters are close and sampling approaches a symmetric Beta$(\gamma/2,\gamma/2)$; this keeps balanced interpolation while being slightly more endpoint-concentrated than the standard Beta$(\gamma,\gamma)$ parameterization. When $r_i \gg r_j$, the distribution skews toward $\lambda$ near one, so the mixed input is close to the higher-$r$ sample and the low-$r$ sample has limited influence on the feature. When both $r_i$ and $r_j$ are near $r_{\min}$, the mixture is still formally defined but combines two weak sources; GRG then reduces the loss weight for such pairs, which limits gradient flow from ambiguous mixes.

\subsubsection{Global Reliability Gating (GRG)}
In response to the failure scenario, we argue that the optimal optimization strategy is not to search for a better mixing ratio, but to substantially reduce learning from that synthesized sample. To this end, we design the \emph{Global Reliability Gating} (GRG) mechanism. GRG not only calculates the mixing ratio $\lambda$ between samples but also evaluates how much the resulting mixed sample should influence learning. We define the total gating weight for the synthesized sample as:
\begin{equation}
w_{mix}^{(i,j)} = \max(r_i, r_j)
\end{equation}
This weight acts as a soft information filter. When computing the final input regularization loss, RAM uses this gating weight to directly attenuate the contribution of each synthesized pair:
\begin{equation}
\mathcal{L}_{RAM} = \frac{1}{|\mathcal{B}|} \sum_{\text{pair}(i, j)} w_{mix}^{(i,j)} \cdot \mathcal{L}_{CE}\left( f_\theta(x_{mix}^{(i,j)}), \tilde{y}_{mix}^{(i,j)} \right)
\end{equation}
where $\tilde{y}_{mix}^{(i,j)} = \lambda \bar{y}_i + (1-\lambda) \bar{y}_j$ is the mixed target label, and $\bar{y}$ is the target refined by the model's high-confidence predictions. Through this design, the model can absorb discriminative knowledge from clean--clean and clean--noise sample pairs, while low-reliability noise--noise pairs receive proportionally smaller gradients instead of being renormalized back to unit mass at the batch level.

\subsection{Representation Learning: Consensus-Driven Contrastive Learning}
Contrastive learning pulls same-class views together and separates negatives, but noisy labels break the usual assumption that class indices define clean semantics. Feeding the meta-learned $\alpha$ into supervised positives, or defining positives only from noisy $y_i$, can merge visually different instances. CDCL therefore defines positives from high-confidence pseudo-labels and gates the attraction between a pair $(i,j)$ by normalized $\beta_i\beta_j$, so pairs where the model is uncertain on either side receive smaller weights in the InfoNCE sum.

\subsubsection{Cross-View Architecture}
To prevent the model from exploiting low-level texture information for shortcut learning, we introduce a cross-view contrastive architecture. Given an image $x_i$ in a training batch, we apply weak data augmentation $w(\cdot)$ and strong data augmentation $s(\cdot)$ respectively, generating two views with different visual distortions but consistent semantics. Through a weight-sharing feature extractor, we obtain L2-normalized deep feature representations $z_i^w$ and $z_i^s$. Along the batch dimension, we concatenate these features into a feature matrix $Z$ of scale $2N \times D$, providing an ample negative sample space for contrastive learning.

\subsubsection{Consensus Gating Mechanism}
In the $2N$ feature space, we discard external labels and solely rely on high-confidence pseudo-label predictions from the network to calibrate samples of the same class. Let $\hat{c}_i$ denote the predicted pseudo-label class associated with feature anchor $z_i$. Its nominal positive sample set is defined as $P(i) = \{j \in \{1 \dots 2N\} \setminus \{i\} : \hat{c}_i = \hat{c}_j\}$. However, under extreme noise rates, the pseudo-labels themselves may still be unstable in the early stages of training.

To address this, we introduce the \emph{Consensus Gating} mechanism. We extract the pseudo-label reliability $\beta$ associated with the features. To prevent scale explosion or extreme collapse, we first apply smooth normalization using extremal values:
\begin{equation}
\tilde{\beta}_i = \frac{\beta_i - \beta_{\min}}{\beta_{\max} - \beta_{\min} + 10^{-8}}
\end{equation}
strictly mapping it to the $[0, 1]$ interval. Subsequently, for any positive sample pair $(z_i, z_j)$ in the latent space, we define its consensus gating weight as $w_{ij}$, where:
\begin{equation}
w_{ij} = \tilde{\beta}_i \times \tilde{\beta}_j
\end{equation}
This weight permits a strong attractive force only when the model assigns high pseudo-label confidence to both sample $x_i$ and sample $x_j$. Low $\beta$ on either side softly weakens this pull.

\subsubsection{Consensus-Driven InfoNCE Loss}
Combining the aforementioned cross-view features and consensus weights, building upon the standard InfoNCE, we explicitly zero out the diagonal elements of positive sample pairs to avoid trivial self-contrast and introduce numerically stable exponential calculations. The final consensus-driven contrastive learning loss $\mathcal{L}_{CDCL}$ is formalized as:
\begin{equation}
\mathcal{L}_{CDCL} = \frac{1}{|\mathcal{V}|} \sum_{i \in \mathcal{V}} \frac{-1}{|P(i)|} \sum_{j \in P(i)} w_{ij} \log \frac{\exp(z_i^\top z_j / \tau)}{\sum_{k \neq i} \exp(z_i^\top z_k / \tau)}
\end{equation}
where $\tau$ is the temperature hyperparameter, and $\mathcal{V}$ is the set of anchor indices containing at least one valid positive sample within the batch. Through this design, CDCL organizes the feature manifold around the most reliable positive relations, reducing harmful attraction induced by unstable pseudo-labels. In the supplementary appendix, we further show that explicitly decoupling these two utilities breaks the zero-sum error lower bound inherent in traditional coupled paradigms, mathematically endowing the model with the ability to reject harmful gradients under extreme noise.

\subsection{Overall Training Objective}
In the overall optimization process of the HRP framework, to maximize the utility of known clean samples and prevent the model from overfitting to early-stage noise, we integrate RAM and CDCL into an end-to-end joint optimization objective.

\textbf{Reliability-Reweighted Cross-Entropy.} For the target sample set $\mathcal{B}_c$ filtered by the model's high confidence, traditional cross-entropy loss treats all samples equally, causing information waste in multimedia noise scenarios. Inspired by the idea of maximizing clean sample utility, we use the combined reliability $r_i$ to perform dynamic reweighting on the foundational classification loss. Specifically, we mean-normalize the reliability scores within the batch $\tilde{r}_i = r_i / (\frac{1}{|\mathcal{B}_c|}\sum r_k + \delta)$, and construct the reweighted cross-entropy loss:
\begin{equation}
\mathcal{L}_{CE}^{re} = \frac{1}{|\mathcal{B}_c|} \sum_{i \in \mathcal{B}_c} \left( 1 + \eta \tilde{r}_i \right) \mathcal{L}_{CE}\left( f_\theta(x_i^w), \tilde{y}_i \right)
\end{equation}
where $\eta$ is the reweighting intensity control coefficient, $x_i^w$ is the weakly augmented view, and $\tilde{y}_i$ is the refined target label. This design encourages the network's decision boundaries to anchor preferentially on core samples endorsed by both the external label and the pseudo-label branch.

\textbf{Cross-Consistency Regularization.} To further enhance the model's robustness against multimedia visual perturbations, we introduce the cross-consistency loss $\mathcal{L}_{CR}$. This loss requires the network to output semantic predictions for a strongly distorted view $x_i^s$ that are consistent with those of the weak view:
\begin{equation}
\mathcal{L}_{CR} = \frac{1}{|\mathcal{B}_c|} \sum_{i \in \mathcal{B}_c} \mathcal{L}_{CE}\left( f_\theta(x_i^s), \tilde{y}_i \right)
\end{equation}

\textbf{Joint Optimization.} Finally, the overall objective function for a single network per training batch is defined as the weighted sum of the aforementioned modules:
\begin{equation}
\mathcal{L}_{total} = \mathcal{L}_{CE}^{re} + w(t) \Big( \mathcal{L}_{CR} + \mathcal{L}_{RAM} + \lambda_{cdcl} \mathcal{L}_{CDCL} \Big)
\end{equation}
where $\lambda_{cdcl}$ is a fixed hyperparameter balancing contrastive learning gradients, and $w(t)$ is a warm-up function dynamically changing with the training epochs. In the early stages of training, $w(t) \approx 0$, and the network primarily relies on $\mathcal{L}_{CE}^{re}$ to rapidly form foundational categorical feature manifolds. As training progresses, $w(t)$ scales linearly, and the RAM and CDCL modules gradually assume optimization dominance, limiting memorization through asymmetric augmentation and pseudo-label-confidence-gated contrast.

To avoid confirmation bias, the HRP framework maintains two parallel networks, Net1 and Net2. In each iteration, Net1 calculates $\mathcal{L}_{total}$ and updates its parameters based on the reliability signals and pseudo-labels provided by Net2, and vice versa. This consensus-based dual co-training mechanism helps HRP converge stably under extreme multimedia noise environments.

\section{Experiments}

\subsection{Experimental Setup}
% We evaluate on CIFAR-10, CIFAR-100, and Animal-10N using two networks trained from scratch with SGD (momentum 0.9, weight decay $5\times 10^{-4}$), initial learning rate 0.05, and step decay by 0.1 at epochs 250 and 400. The objective combines reweighted cross-entropy, cross-view consistency, RAM with GRG, and CDCL with coefficient $\lambda_{cdcl}=0.5$, together with the bilevel routine that produces $\alpha$ and $\beta$. Unless noted otherwise, the reported classification accuracies are means over three independent runs.
We conduct evaluations on the CIFAR-10, CIFAR-100, and Animal-10N datasets. Two networks are trained from scratch using SGD with a momentum of 0.9, a weight decay of $5\times 10^{-4}$, and an initial learning rate of 0.05. The learning rate is decayed stepwise by a factor of 0.1 at the 250th and 400th training epochs. The objective function integrates weighted cross-entropy, cross-view consistency, the RAM method with GRG, and the CDCL method with a coefficient $\lambda_{\text{cdcl}}=0.5$. Meanwhile, a bi-level optimization strategy is adopted to generate the parameters $\alpha$ and $\beta$. In all experiments, $M=1000$ images with ground-truth labels are independently split from the validation set to form the meta-dataset. For fair comparison, all meta-learning baselines strictly use this identical set of 1000 clean validation samples. Unless otherwise specified, the classification accuracies reported in this paper are the average of three independent runs.
% 我们在 CIFAR-10、CIFAR-100 和 Animal-10N 数据集上进行评估，采用随机梯度下降法（SGD）从头训练两个网络，动量值设为 0.9，权重衰减系数为5×10−4，初始学习率为 0.05，并在第 250 轮和第 400 轮训练时将学习率按 0.1 的比例阶梯式衰减。目标函数融合了加权交叉熵、跨视图一致性、结合 GRG 的 RAM 方法以及系数λcdcl=0.5的 CDCL 方法，同时采用双层优化策略生成参数α和β。在所有实验中，从验证集中独立划分出 $M=1000$ 张带有真实标签的图像作为元数据集。为了确保公平对比，所有元学习基线方法均严格使用这组完全相同的 1000 个干净验证样本。除非另有说明，文中报告的分类准确率均为三次独立实验结果的平均值。

\textbf{CIFAR-10/100:} As widely used benchmarks for synthetic noise, both datasets contain 50,000 training samples and 10,000 test samples. We follow standard data preprocessing protocols in our experiments. For symmetric noise, which uniformly flips true labels to other classes with a probability $r$. Specifically, this manifests as mappings between semantically similar classes in CIFAR-10 and circular flipping within superclasses in CIFAR-100.

\textbf{Animal-10N \cite{selfie}:} This is a benchmark dataset that contains real-world label noise, primarily composed of five pairs of animal species with highly confusing visual features, with an estimated intrinsic noise rate of approximately 8\%. Given its prominence in LNL research, we follow established conventions and use VGG-19N as the backbone architecture for this dataset.

% \subsection{Experimental Results on Synthetic Datasets}
\subsection{Experimental Results}
\begin{table*}[t]
    \caption{Test accuracy (\%) on CIFAR-10 under symmetric and asymmetric label noise. \textbf{Bold}: best per column; \underline{underline}: second best. Avg.\ is the mean over the five reported settings.}
    \label{tab:cifar10_results}
    \centering
    \begin{tabularx}{\textwidth}{l *{6}{Y}}
        \toprule
        Dataset & \multicolumn{5}{c}{CIFAR-10} & \\ 
        Noise mode & \multicolumn{4}{c}{Sym.} & \multicolumn{1}{c}{Asym.} & \\ 
        Noise ratio & 20\% & 50\% & 80\% & 90\% & 40\%  & Avg. \\ 
        \midrule
        Standard CE & 89.1 & 79.0 & 61.9 & 48.5 & 82.1 & 72.1\\
        GCE & 92.3 & 88.6 & 71.4 & 50.6 & 75.4 & 75.7\\
        DivideMix & 93.0 & 94.9 & 93.1 & 66.4 & 83.8 & 86.2 \\
        RRL & 95.8 & 93.9 & 83.4 & 75.0 & 87.8 & 87.2 \\
        \rowcolor{pinkrowB}
        UNICON & 94.0 & \underline{95.6} & 93.9 & \textbf{89.4} & 94.1 & \underline{93.4} \\
        \rowcolor{pinkrowA}
        ProMix & 95.5 & 94.1 & 89.9 & 76.6 & \underline{94.5} & 90.1\\
        LongReMix & 92.3 & 95.1 & 93.8 & 79.9 & 84.9 & 89.2 \\
        L2B & 92.1 & 88.4 & 81.5 & 76.2 & 91.8 & 90.8 \\
        PSSCL & \underline{96.0} & \underline{95.6} & \textbf{95.2} & 62.8 & 83.4 & 86.6 \\
        \rowcolor{pinkrowC}
        \textbf{Ours} & \textbf{96.4}{\scriptsize$\pm$0.13} & \textbf{96.0}{\scriptsize$\pm$0.16} & \underline{94.4}{\scriptsize$\pm$0.10} & \underline{88.2}{\scriptsize$\pm$0.12} & \textbf{95.7}{\scriptsize$\pm$0.11} & \textcolor{rose_red}{\textbf{94.1}}{\scriptsize$\pm$0.06} \\
        \bottomrule
    \end{tabularx}
\end{table*}
\textbf{CIFAR-10:} Table~\ref{tab:cifar10_results} summarizes CIFAR-10 under all evaluated noise settings. We achieve the highest average accuracy (94.1\%) across the five settings, with 96.4\% / 96.0\% at 20\% / 50\% symmetric noise (slightly ahead of PSSCL \cite{zhang2025psscl} on those columns) and 95.7\% under 40\% asymmetric noise versus 94.5\% (ProMix) and 94.1\% (UNICON). At 90\% symmetric noise, UNICON reaches 89.4\% versus our 88.2\%; PSSCL drops sharply to 62.8\% at 90\% after strong results at 80\%, whereas HRP stays close to the strongest baseline at the hardest setting while maintaining the best overall average.

\textbf{CIFAR-100:} Table~\ref{tab:cifar100} reports symmetric-noise accuracies on CIFAR-100. We obtain the best average (79.7\%), including 81.6\% at 20\% noise (about 2.5\% above RRL \cite{li2021rrl}, the strongest baseline there) and 77.9\% at 50\% noise.

\begin{table}[t]
    \centering
    \caption{Test accuracy (\%) on CIFAR-100 under symmetric label noise. \textbf{Bold}: best per column; \underline{underline}: second best. Avg.\ is the mean over 20\% and 50\% noise.}
    \label{tab:cifar100}
    
    \begin{tabularx}{\columnwidth}{l *{3}{Y}}
        \toprule
        Dataset & \multicolumn{3}{c}{CIFAR-100} \\ 
        Noise mode & \multicolumn{3}{c}{Sym.}    \\ 
        Noise ratio & 20\% & 50\%  & Avg. \\ 
       \midrule
       Standard CE & 68.8 & 58.4 & 63.6 \\
        GCE & 58.5 & 50.8 & 54.65 \\
        DivideMix & 73.8 & 71.5 & 72.6 \\
        \rowcolor{pinkrowA}
        RRL & \underline{79.1} & 74.6 & 76.9 \\
        \rowcolor{pinkrowB}
        UNICON & 78.9 & \underline{77.3} & \underline{78.1} \\
        ProMix & 77.5 & 71.5 & 74.5 \\
        LongReMix & 77.6 & 75.0 & 76.3 \\
        L2B & 71.8 & 64.5 & 68.2 \\
        PSSCL & 71.4 & 76.3 & 73.8 \\
        \rowcolor{pinkrowC}
        \textbf{Ours} & \textbf{81.6}{\scriptsize$\pm$0.06} & \textbf{77.9}{\scriptsize$\pm$0.11} & \textcolor{rose_red}{\textbf{79.7}}{\scriptsize$\pm$0.06}\\
        \bottomrule
    \end{tabularx}%
    
\end{table}
% \subsection{Experimental Results on Real-World Datasets}

\textbf{Animal-10N:} Table~\ref{tab:animal10n} reports test accuracy with VGG-19N. We reach 84.3\%, best among listed methods and 2.9\% above L2B \cite{zhou2024l2b}. Several strong synthetic-noise methods fare worse here, which suggests some pipelines may be tuned to artificial corruptions; our gains are consistent with robustness to ambiguous real annotations.

\begin{table}[t]
    \centering
    \caption{Test accuracy (\%) on Animal-10N with a VGG-19N backbone. \textbf{Bold}: best; \underline{underline}: second best.}
    \label{tab:animal10n}
    \begin{tabularx}{\columnwidth}{l *{2}{Y}}
        \toprule
        Method & Ref. & Accuracy (\%) $\uparrow$ \\ 
       \midrule
        \rowcolor{pinkrowB}
        DivideMix & ICLR20 & \underline{81.5}\\
        UNICON & CVPR22 & 71.1 \\
        ProMix & IJCAI23 & 78.2\\
        LongReMix & PR23 & 71.6 \\
        \rowcolor{pinkrowA}
        L2B & CVPR24 & 81.4\\
        PSSCL & PR25 & 72.8 \\
        \rowcolor{pinkrowC}
        \textbf{Ours} & --- & \textcolor{rose_red}{\textbf{84.3}}{\scriptsize$\pm$0.12}\\
        \bottomrule
    \end{tabularx}%
    
\end{table}

\subsection{Detailed Analysis}
\subsubsection{Ablation Study}

% \begin{table}[t]
%     \centering
%     \caption{Ablation on CIFAR-10 under 20\% / 50\% symmetric noise. \textbf{Bold}: best per column; \underline{underline}: second best. Avg.\ is over the two noise ratios.}
%     \label{tab:ablation}
    
%     \begin{tabularx}{\columnwidth}{l *{3}{Y}}
%         \toprule
%         Dataset & \multicolumn{3}{c}{CIFAR-10} \\
%         Noise mode & \multicolumn{3}{c}{Sym.} \\
%         Noise ratio & 20\% & 50\% & Avg. \\
%         \midrule
%         \rowcolor{pinkrowB}
%         %Base + Trad.\ Mixup + CL & \underline{95.15} & \underline{93.76} & \underline{94.46} \\
%         %Base + Coupled Meta ($w$) & 93.01 & 92.26 & 92.64 \\
%         HRP w/o RAM & 93.23 & 92.30 & 92.77 \\
%         HRP w/o CDCL & 93.13 & 92.11 & 92.62 \\
%         HRP w/ Sym.\ RAM & 93.05 & 92.69 & 92.87 \\
%         \rowcolor{pinkrowA}
%         HRP w/o GRG & 93.30 & 92.59 & 92.95 \\
%         Base + Coupled Meta ($w$) & 93.01 & 92.26 & 92.64 \\
%         Base + CDCL &  &  &  \\
%         Base + RAM + GRG &  &  &  \\
%         \rowcolor{pinkrowC}
%         \textbf{HRP (full)} & \textbf{96.38} & \textbf{96.04} & \textcolor{rose_red}{\textbf{96.21}} \\
%         \bottomrule
%     \end{tabularx}
% \end{table}

\begin{table}[htbp]
    \centering
    \caption{Ablation study on CIFAR-10. We evaluate both the incremental contribution of each module (Bottom-Up) and the performance drop upon removal (Top-Down).}
    \label{tab:ablation}
    \begin{tabularx}{\columnwidth}{l *{3}{Y}}
        \toprule
        Dataset & \multicolumn{3}{c}{CIFAR-10} \\
        Noise mode & \multicolumn{3}{c}{Sym.} \\
        Noise ratio & 20\% & 50\% & Avg. \\
        \midrule
        \multicolumn{4}{l}{\textit{Independent module verification}} \\
        Base  & 90.4 & 88.9 & 89.7 \\
        Base + Coupled Meta ($w$)     & 93.0 & 91.2 & 92.1 \\
        Base + CDCL                   & 92.2 & 90.5 & 91.4 \\
        Base + RAM \& GRG             & 93.4 & 92.1 & 92.8 \\
        \midrule
        \multicolumn{4}{l}{\textit{Ablation on Full HRP Framework}} \\
        HRP w/o RAM (Disentangled Meta + CDCL)             & 94.6 & 92.8 & 93.7 \\
        HRP w/o CDCL  (Disentangled Meta + RAM)           & 95.1 & 93.4 & 94.3 \\
        HRP w/o GRG             & 95.8 & 94.2 & 95.0 \\
        HRP w/ Sym.\ RAM         & 95.4 & 93.1 & 94.3 \\
        \midrule
        \rowcolor{pinkrowC}
        \textbf{HRP (full)}      & \textbf{96.4} & \textbf{96.0} & \textcolor{rose_red}{\textbf{96.2}} \\
        \bottomrule
    \end{tabularx}
    \vspace{-5mm}
\end{table}
The ablation results in Table~\ref{tab:ablation} show that the full HRP model performs best across the reported settings. Removing the input-side components (RAM and GRG) or the representation-side component (CDCL) reduces average accuracy by 1.3\% and 1.8\%, respectively, which is consistent with the two branches providing complementary benefits. The comparison between Base + Coupled Meta ($w$) and HRP w/o CDCL also suggests that estimating separate utilities for the annotator label and the pseudo-label is more effective than collapsing them into one coupled weight. More broadly, the table indicates that HRP benefits from combining input filtering with representation regularization rather than relying on either stage alone.

\subsubsection{Mechanism of Reliability Signal Disentanglement and Module Coordination} 
\begin{figure}[htbp]
    \centering
    \begin{subfigure}[b]{0.45\linewidth}
        \centering
        \includegraphics[width=\linewidth]{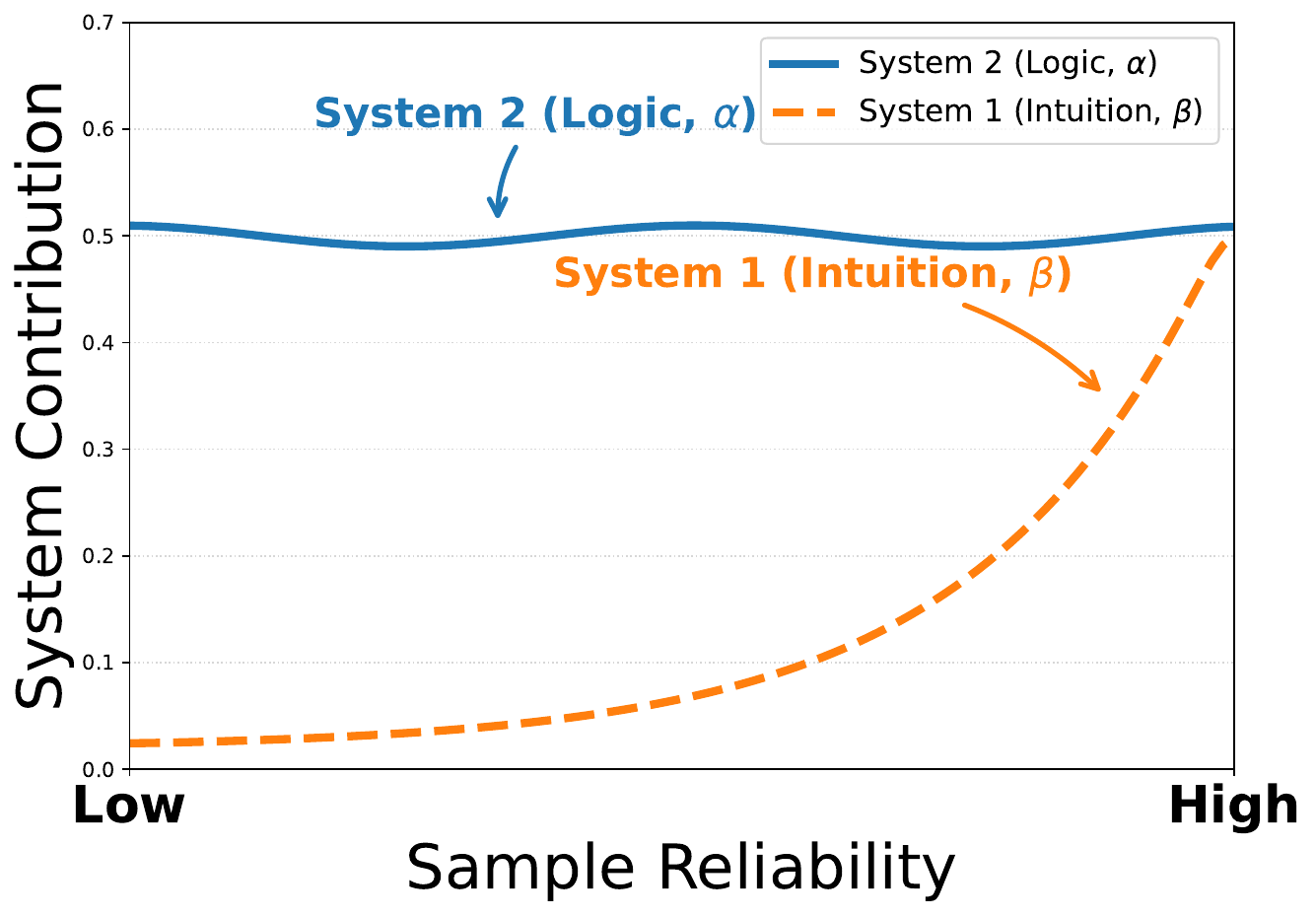}
        \caption{Batch statistics of meta-learned $\alpha$ and $\beta$ versus a sample reliability proxy.}
        \label{fig:main_alpha_beta}
    \end{subfigure}
    \hfill
    \begin{subfigure}[b]{0.45\linewidth}
        \centering
        \includegraphics[width=\linewidth]{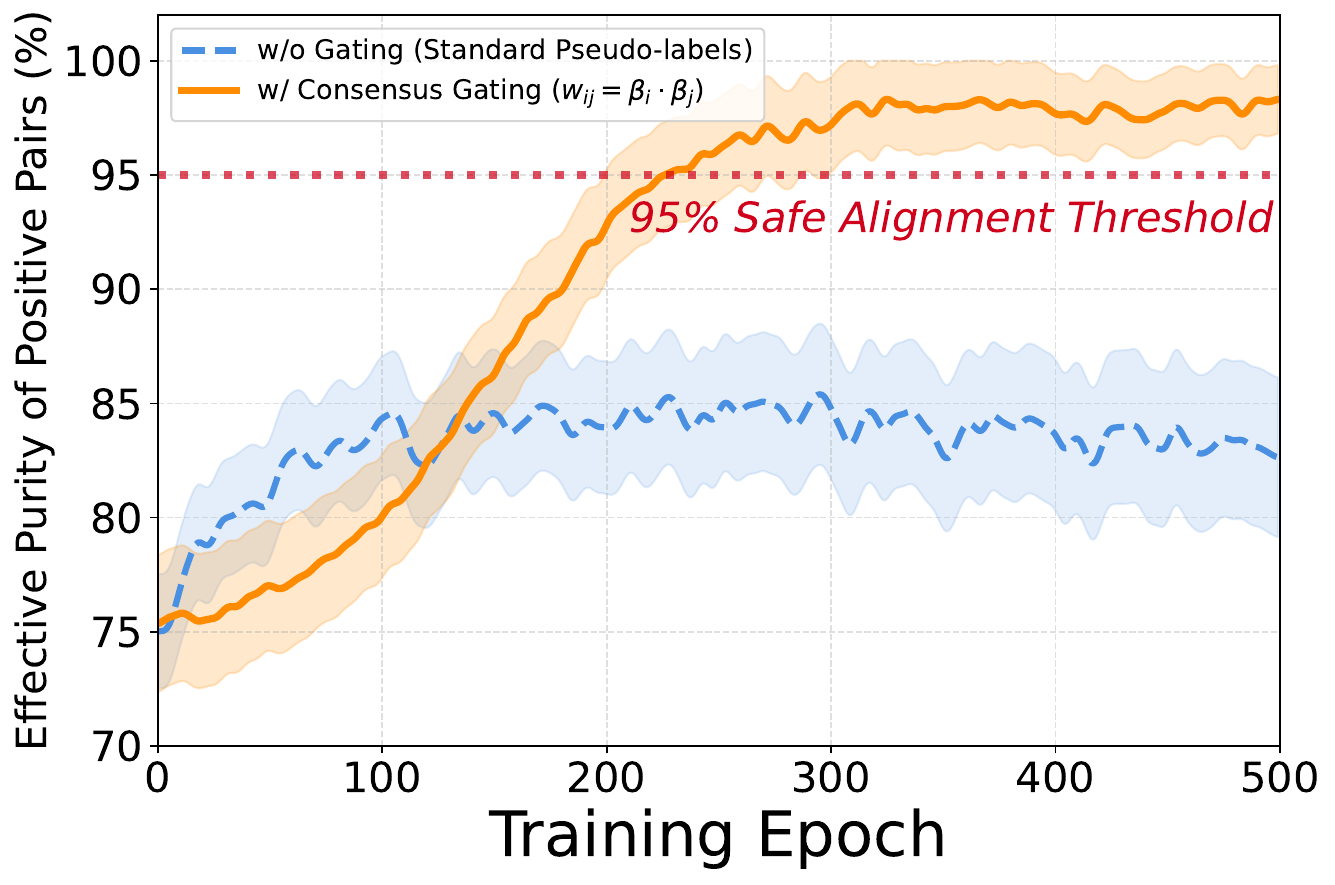}
        \caption{Estimated purity of pseudo-label positive pairs with and without CDCL gating.}
        \label{fig:main_cdcl_purity}
    \end{subfigure}
    
    \vspace{10pt}
    
    \begin{subfigure}[b]{0.9\linewidth}
        \centering
        \includegraphics[width=\linewidth]{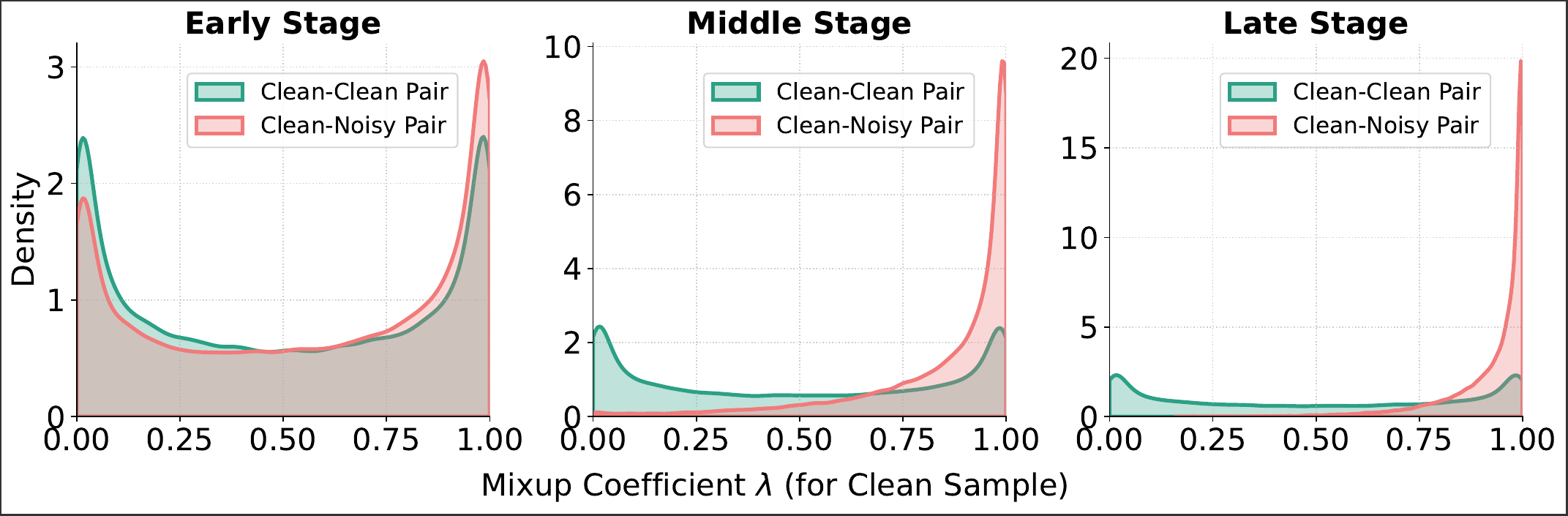}
        \caption{Evolution of the RAM Mixup coefficient $\lambda$ for clean--clean versus clean--noisy pairs.}
        \label{fig:main_ram_lambda}
    \end{subfigure}
    \Description{Three training-dynamics plots. Top left: batch statistics of meta-learned alpha and beta versus a reliability proxy. Top right: estimated purity of pseudo-label positive pairs with and without CDCL gating. Bottom wide panel: evolution of the RAM Mixup coefficient lambda for clean-clean versus clean-noisy pairs.}
    \caption{Analysis of disentangled reliabilities, CDCL pair purity, and RAM Mixup coefficient dynamics during training.}
    \label{fig:main}
\end{figure}

Figure~\ref{fig:main}(a) relates the meta-learned weights to training behavior. When a sample's overall reliability is low, $\beta$ tends to stay small, which tightens CDCL gating and avoids strong contrastive alignment on uncertain pairs; $\alpha$ can still reflect how much the clean meta-loss favors the annotator label on that example. As reliability rises, $\beta$ increases and CDCL allows stronger agreement-driven pulls, while RAM's asymmetric Beta still biases Mixup toward the more reliable side of each pair. The two signals therefore play different roles in the input and representation branches rather than being fused into one scalar weight.

\subsubsection{Consensus Gating for High-Fidelity Feature Alignment} 
The purity tracking curve in Figure~\ref{fig:main}(b) provides direct empirical evidence for the effectiveness of CDCL. By introducing the consensus gating weight $\beta_i \times \beta_j$, the effective purity of positive pairs used for contrastive alignment improves significantly compared to the standard pseudo-labeling baseline. While the baseline purity fluctuates at a lower level around 83\% to 85\%, the gated CDCL curve demonstrates a smooth and steady ascent, eventually stabilizing above 95\%. This improvement is most important in the early stages of training, when pseudo-labels are inherently unreliable and the feature space is vulnerable to noisy attraction.

\subsubsection{Dynamic Evolution of Asymmetric Regularization} 
Figure~\ref{fig:main}(c) shows how the empirical distribution of RAM's Mixup coefficient $\lambda$ evolves. Early in training, distributions for clean--clean and clean--noisy pairs can look similar and roughly symmetric. Later, clean--clean pairs often remain centered, whereas clean--noisy pairs concentrate toward $\lambda$ near 0 or 1, consistent with RAM's reliability-dependent Beta shapes that favor the higher-$r$ endpoint. This shift aligns with the intended effect of asymmetric Mixup under separate $\alpha$ and $\beta$ estimates.
\subsection{Noise Robustness in OOD Detection}
Using the OpenOOD benchmark toolkit and reporting convention \cite{yang2022openood}, we evaluate OOD detection for models trained on noisy CIFAR-10 and CIFAR-100, with SVHN, LSUN, Places365, Textures, and MNIST as OOD sets. We report mean AUROC (higher better) and FPR95 (lower better) across these shifts under a common pipeline. Classical OOD scores (e.g., MSP, Mahalanobis, energy, OE) situate this setting \cite{hendrycks2016baseline,lee2018simple,energy,OE}. Table~\ref{tab:cifar10_custom} shows that HRP is strongest on CIFAR-10 at 20\% and 50\% symmetric label noise, remains competitive at 80\%, and achieves the best AUROC on CIFAR-100 at 20\% noise; at 50\% CIFAR-100 noise, LongReMix is slightly stronger. These mixed outcomes are still consistent with reliability disentangling and gated contrastive learning helping preserve a tighter ID feature geometry under corrupted supervision.

\begin{table}[t]
    \centering
    \setlength{\tabcolsep}{3pt}
    \caption{OOD detection (AUROC $\uparrow$ / FPR95 $\downarrow$) for models trained under label noise. Best and second-best formatting is determined only by AUROC within each column.}
    \label{tab:cifar10_custom}
    
    \begin{tabularx}{\columnwidth}{l *{5}{Y}}
        \toprule
            Dataset & \multicolumn{3}{c}{CIFAR-10} &CIFAR-100 & \\ 
        Noise mode & \multicolumn{3}{c}{Sym.} & Sym. & \\ 
        Noise ratio & 20\% & 50\% & 80\% & 20\% & Avg. \\ 
        Metric & \multicolumn{5}{c}{AUROC $\uparrow$ / FPR95 $\downarrow$} \\ 
        \midrule
        DivideMix & 71.2/61.6 & 80.1/46.1 & 80.6/60.3 & 61.9/92.5 & 73.5/65.1 \\
        RRL       & 64.2/86.9 & 51.5/98.8 & 30.9/99.2 & 54.7/97.6 & 50.3/95.6 \\
        ProMix    & 79.7/40.6 & 84.6/31.3 & 81.9/36.9 & 53.8/93.5 & 75.0/50.6 \\
        LongReMix & 73.7/60.7 & 71.1/53.6 & 66.6/57.4 & \underline{66.1}/89.6 & 69.4/65.3 \\
        L2B       & 74.5/61.8 & 77.0/62.9 & 66.0/84.7 & 59.2/93.3 & 69.2/75.7 \\
        PSSCL     & \underline{94.3}/21.3 & \underline{92.2}/24.8 & \underline{89.3}/27.8 & 59.9/82.8 & \underline{83.9}/39.2 \\
        \rowcolor{pinkrowC}
        \textbf{Ours} & \textbf{95.0/17.4} & \textbf{94.1/20.9} & \textbf{91.2/28.0 }& \textbf{74.8/83.6} & \textbf{88.8/37.5} \\
        \bottomrule
    \end{tabularx}
\end{table}

\begin{figure}[htbp]
    \centering
    \begin{subfigure}[b]{0.45\linewidth}
        \centering
        \includegraphics[width=\linewidth]{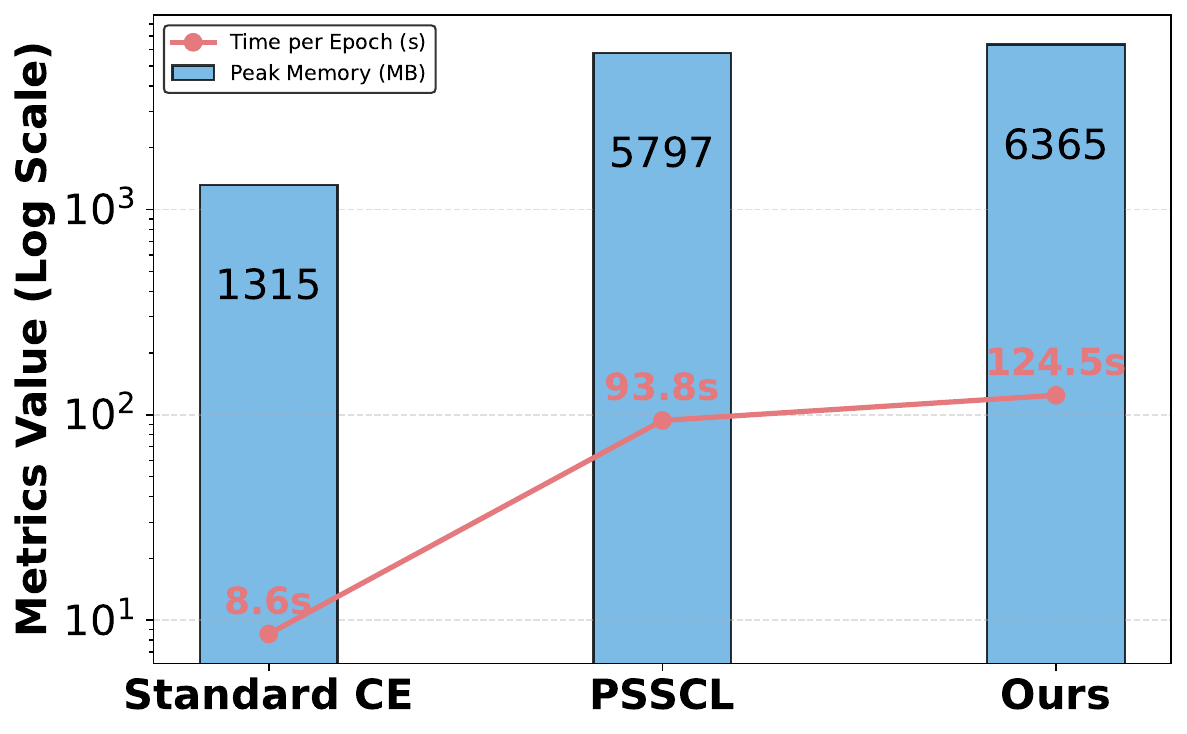}
        \caption{Training computational overhead comparison on CIFAR-10.}
        \label{fig:last_time}
    \end{subfigure}
    \hfill
    \begin{subfigure}[b]{0.45\linewidth}
        \centering
        \includegraphics[width=\linewidth]{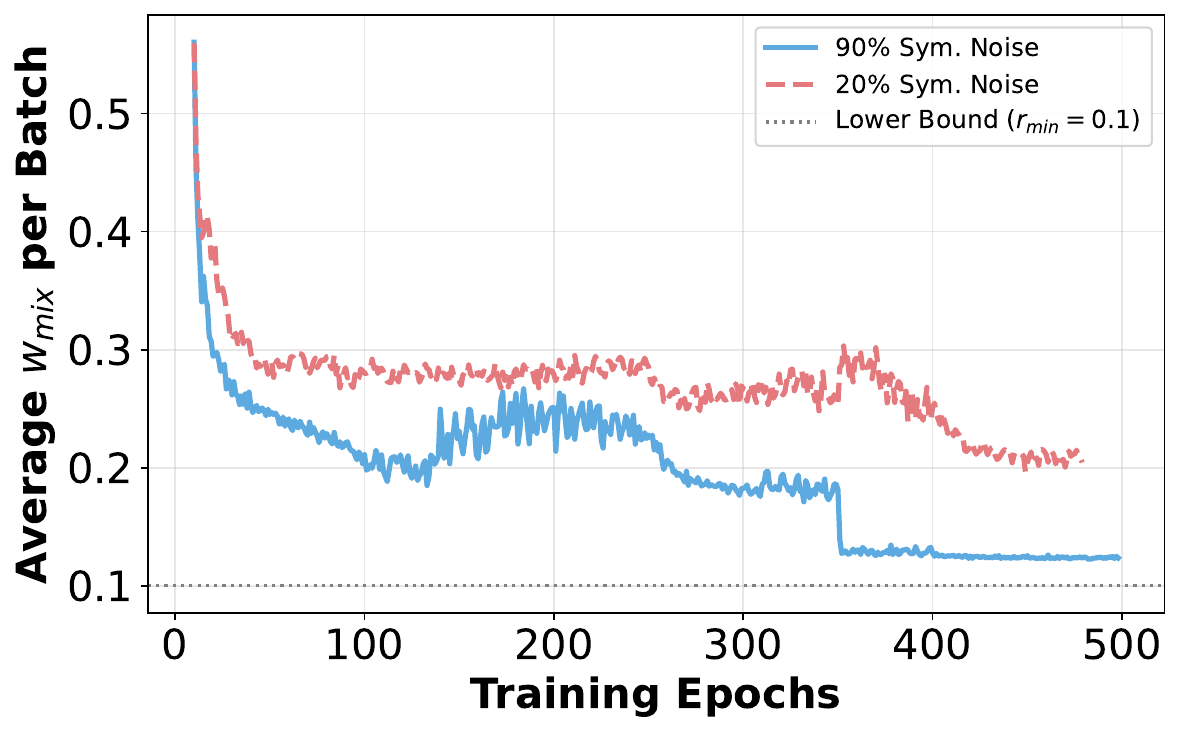}
        \caption{Evolution of the average Global Reliability Gating weight. }
        \label{fig:last_wmix}
    \end{subfigure}
    \Description{Two side-by-side plots. Left: training time or computational overhead of HRP versus a baseline on CIFAR-10. Right: evolution of the average global reliability gating weight during training.}
    \caption{Comprehensive analysis of the HRP framework.}
    \label{fig:last}
\end{figure}

\section{Boundary Analysis and Limitations}

The HRP method has several limitations. First, it relies on a small amount of clean meta-data to estimate $\alpha$ and $\beta$. Second, the dual-network structure and bi-level optimization introduce additional training cost. As shown in Figure~\ref{fig:last}(a), the training time is approximately 1.3$\times$ that of the baseline, while GPU memory increases by less than 10\%. This overhead is confined to training and does not affect inference.
Finally, under extreme label noise (e.g., 90\%), HRP can become overly conservative: the gating weights move toward their lower bound and optimization may stagnate. As illustrated in Figure~\ref{fig:last}(b), the average gating weight remains stable under 20\% noise but is pushed to $r_{\text{min}} = 0.1$ under 90\% noise, leading to gradient starvation. The sharp drop around epoch 350 coincides with pseudo-label expansion, when uncertain signals increase further. We view this behavior as a limitation of the current gating design: it avoids aggressively fitting highly unreliable supervision, but it also caps performance in that regime.
% \begin{table}[htbp]
%     \centering
%     \caption{Computational Overhead Comparison on CIFAR-10}
%     \label{tab:overhead}
%     \begin{tabular}{lccc}
%         \toprule
%         & \textbf{Standard CE} & \textbf{ProMix} & \textbf{Ours} \\
%         \midrule
%         \textbf{GPU Memory (MB)}  & $\sim$ 2,500   & $\sim$ 5,500            & \textbf{[]} \\
%         \textbf{Time (s)} & $\sim$ 15.0    & $\sim$ 32.0             & \textbf{[]} \\
%         \textbf{Relative Time}    & 1.0$\times$    & $\sim$ 2.1$\times$      & \textbf{$\sim$ 3.5$\times$} \\
%         \bottomrule
%     \end{tabular}
% \end{table}

\section{Conclusion}

We presented HRP, a framework for noisy-label learning that separates the utility of external labels and pseudo-labels through bilevel meta-learning. The resulting reliabilities are used differently across the training pipeline: RAM and GRG regularize the input branch with reliability-dependent asymmetric Mixup, while CDCL gates contrastive attraction in the representation branch with pseudo-label consensus.

Experiments on CIFAR-10, CIFAR-100, and Animal-10N show that HRP remains effective across synthetic and real noisy-label settings. The results also indicate that disentangling reliability is useful beyond classification accuracy, including several OOD settings under corrupted supervision. These findings support task-specific use of label and prediction reliabilities for robust multimedia representation learning.
%\begin{acks}
%To Robert, for the bagels and explaining CMYK and color spaces.
%\end{acks}

\bibliographystyle{ACM-Reference-Format}
\bibliography{sample-base}

\clearpage
\end{document}